\documentclass[11pt]{compstvg}

\usepackage{latexsym}

\ifPDFTeX
  \usepackage[T1]{fontenc}
\fi

\ifPDFTeX
  \usepackage[utf8]{inputenc}
\fi

\usepackage{microtype}

\ifPDFTeX
  \usepackage{inconsolata}
\fi

\usepackage{graphicx}
\graphicspath{{./}{../}}
\usepackage{amsmath}
\usepackage{amsthm}
\usepackage{amssymb}
\usepackage{booktabs}
\usepackage{multirow}
\usepackage{array}
\usepackage{adjustbox}
\usepackage{colortbl}
\usepackage{xcolor}
\usepackage{algorithm}
\usepackage{algpseudocode}
\usepackage[most]{tcolorbox}
\usepackage{placeins}
\definecolor{dividergray}{RGB}{240,240,240}
\definecolor{ourslavender}{HTML}{E3F2EF}
\definecolor{myblue}{HTML}{087F7A}
\newcommand{\best}[1]{\textbf{#1}}
\newtheorem{definition}{Definition}
\newtcolorbox[auto counter, number within=section]{promptbox}[2][]{%
  colback=xiaoxibg,
  colframe=xiaoxiblue,
  width=\linewidth,
  arc=2mm,
  boxrule=0.5mm,
  title={\normalsize\raisebox{-0.15ex}{\includegraphics[height=0.9em]{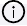}}\hspace{0.5em}#2},
  breakable=true,
  fonttitle=\sffamily\bfseries\large,
  fontupper=\small,
  drop shadow southeast,
  left=5pt, right=5pt,
  top=2pt, bottom=2pt,
  box align=top,
  before=\par\noindent,
  after=\par\vspace{3pt},
  #1
}

\title{Learning \textcolor{tealcomp}{Comp}ositional \textcolor{teals}{S}patio-\textcolor{tealt}{T}emporal \textcolor{tealv}{V}ideo \textcolor{tealg}{G}rounding with Synthetic Curriculum}

\author{\textbf{Xingjian Wang}\textsuperscript{1, 3\ensuremath{*}}, \textbf{Shijian Wang}\textsuperscript{1,2,\ensuremath{*},\ensuremath{\dagger}}, \textbf{Yibo Wang}\textsuperscript{1,2}, \textbf{Zihao Yu}\textsuperscript{3} \\
  \textbf{Runhao Fu}\textsuperscript{1}, \textbf{Xuelian Cheng}\textsuperscript{1,\ensuremath{\ddagger}}, \textbf{Zongyuan Ge}\textsuperscript{1} \\
  \textnormal{\textsuperscript{1}Monash University; \textsuperscript{2}Southeast University; \textsuperscript{3}Shanghai University of Electric Power} \\
  \textnormal{\textsuperscript{\ensuremath{*}}Equal contribution; \textsuperscript{\ensuremath{\dagger}}Project Leader; \textsuperscript{\ensuremath{\ddagger}}Corresponding authors}}
\contact{\raisebox{-0.15ex}{\includegraphics[height=0.9em]{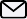}}\hspace{0.4em}Contact: \texttt{themaoqiu@gmail.com}}

\begin{document}
\maketitle
\begin{abstract}

Despite the impressive progress of recent MLLMs on spatio-temporal video grounding (STVG), existing evaluations and training data focus primarily on simple queries. They largely overlook the compositional queries prevalent in real-world scenarios, where a target must be disambiguated by jointly reasoning about its attributes and relations to other entities. To bridge this gap, we propose Compositional Spatio-Temporal Video Grounding (\textsc{CompSTVG}), a task that requires models to process complex textual queries where every intertwined attribute and relational cue is essential for disambiguation. To facilitate this task at scale, we build a synthetic data engine that leverages a spatio-temporal scene graph as a difficulty measure and casts difficulty-controlled query synthesis as a constraint programming problem, producing difficulty-graded data for both evaluation and training. Built on this engine, we introduce \textsc{STVG-CompBench}, a benchmark stratified by explicit difficulty levels that jointly capture temporal complexity and spatial interference. Evaluating 11 representative STVG models on \textsc{STVG-CompBench} reveals that current models perform poorly on compositional queries, exhibiting a sharp performance drop that is typically obscured by overall dataset-level averages. We further construct synthetic training data and propose \textsc{CurrSTVG}, a curriculum reinforcement learning framework that delivers consistent gains, with the largest improvements observed on the most challenging compositional queries.


\end{abstract}

\section{Introduction}

\begin{figure*}[t]
\centering
\includegraphics[width=\linewidth]{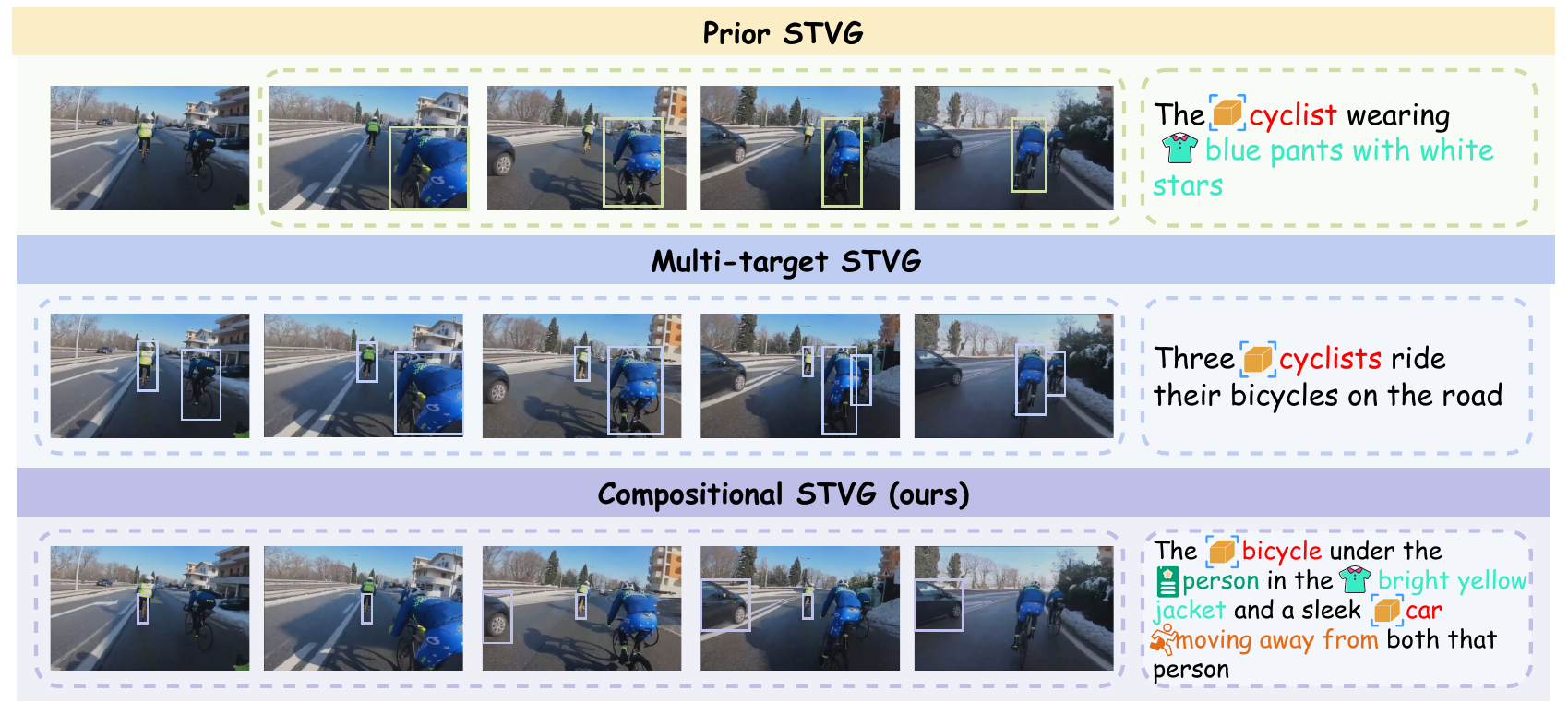}
\caption{Existing STVG vs.\ Compositional STVG. Existing STVG tasks typically pair each tube with a simplistic human-annotated sentence. A real user query intricately composes attribute and relational cues to disambiguate the target against same-category distractors and look-alike time intervals.}
\label{fig:teaser}
\end{figure*}

Spatio-temporal video grounding (STVG) requires models to jointly localize a target entity in both space and time within an untrimmed video given a natural language query~\citep{vidstg}. This capability underpins a wide range of downstream applications, including video retrieval~\citep{frozen,clip4}, video editing~\citep{tokenflow}, complex video reasoning~\citep{vot}, and embodied perception~\citep{robo}. Despite the impressive progress of recent MLLMs on STVG~\citep{llavast,videochat,stvgr1,videomolmo}, prevailing STVG training corpora and benchmarks~\citep{hcstvg,vidstg,yao2025omnistvg} are overwhelmingly populated by simplistic queries that can be resolved using superficial or singular cues. In contrast, real-world user queries frequently demand compositional disambiguation, where the target can only be singled out by jointly reasoning about its attributes and relations to other entities. For instance, locating ``The bicycle under the person in the bright yellow jacket and a sleek car moving away from both that person'' in a scene with multiple cyclist, multiple bicycles, and several riding events requires intricate multi-step reasoning (Figure~\ref{fig:teaser}). Consequently, the ability of current models to perform such compositional reasoning remains a critical vulnerability, largely obscured by existing training and evaluation paradigms.

Although the broader concept of compositionality has been explored in related video tasks, including AGQA~\citep{agqa}, previous research on compositional grounding has primarily evolved along two adjacent axes, though neither directly transfers to our problem context. Compositional Temporal Grounding~\citep{comtg} focuses on novel lexical compositions within the linguistic modality, while Compositional Spatio-Temporal Action Localization~\citep{something,actiongenome,home} focuses on novel action-object combinations within human-object trajectories. Both lines of research fail to account for the rich, multi-faceted nature of visual targets, whereas an STVG query naturally intertwines appearance, action, temporal order, spatial relations, and inter-object interactions within a single sentence. Therefore, a comprehensive task formulation, dataset, and evaluation protocol that explicitly target compositional STVG are still absent from the literature.

To address this gap, we propose \textbf{Comp}ositional \textbf{S}patio-\textbf{T}emporal \textbf{V}ideo \textbf{G}rounding, a novel task challenging models to process complex queries where every cue, drawn from appearance, action, temporal, spatial, and relational attributes, is strictly essential for disambiguation. To facilitate this task at scale, we build a synthetic data engine leveraging a spatio-temporal scene graph (STSG) extracted from each video. We define query difficulty based on the ambiguity introduced by temporal and spatial distractors enumerated on this graph, and select the minimal attribute subset required to disambiguate the target by solving a binary integer programming problem. This engine yields strictly difficulty-graded compositional STVG data suitable for both evaluation and training.

Building upon this engine, we introduce \textsc{STVG-CompBench}, a benchmark stratified across explicit difficulty tiers. Extensive evaluation of eleven representative grounders reveals a severe and consistent performance degradation as query complexity increases. This limitation persists across both transformer-based and MLLM-based paradigms, confirming that the bottleneck stems from a fundamental lack of compositional reasoning capabilities rather than architectural artifacts. Finally, we propose \textsc{CurrSTVG}, a curriculum reinforcement learning framework that feeds these difficulty labels back into the training loop via GRPO. Guided by formatting, temporal, and spatial rewards, our framework schedules training samples from easy to hard, ensuring that the same complexity metric used for benchmarking actively drives model optimization. \textsc{CurrSTVG} delivers consistent performance gains, yielding the most significant improvements on the most challenging, highly intertwined compositional queries.

In summary, our main contributions are three-fold: (i) \textsc{CompSTVG}, a new task and a scalable STSG-based synthetic data engine for compositional STVG; (ii) \textsc{STVG-CompBench}, a difficulty-stratified benchmark exposing an architecture-agnostic compositional bottleneck across $11$ representative grounders; and (iii) \textsc{CurrSTVG}, a curriculum RL framework that converts the same difficulty axis into the largest gains on the hardest compositional queries.

\section{Related Work}
\label{sec:related}

\paragraph{Spatio-Temporal Video Grounding.} Early two-stage methods such as STGRN~\citep{vidstg} and STGVT~\citep{hcstvg} run a pretrained detector to obtain region proposals and learn a matching network to pick the one referred to by the query. Transformer-based one-stage grounders such as TubeDETR~\citep{tubedetr}, CG-STVG~\citep{cgstvg} and TA-STVG~\citep{tastvg} regress the spatio-temporal tube end-to-end but remain bound to a fixed text encoder and degrade on long, multi-clause queries. Recent work turns to MLLMs, where LLaVA-ST~\citep{llavast} and SpaceVLLM~\citep{spacevllm} inject learnable spatio-temporal tokens, VideoChat-R1~\citep{videochat} fine-tunes with multi-task GRPO under an IoU reward, and STVG-R1~\citep{stvgr1} and VideoMolmo~\citep{videomolmo} couple a VLM with SAM2~\citep{sam2}. Existing benchmarks~\citep{vidstg,hcstvg,omniground,yao2025omnistvg} pool queries from one obvious cue with queries demanding multi-attribute composition, hiding the regime our benchmark probes. \textsc{CurrSTVG} departs from these methods on the data axis, extracting a difficulty signal from the data itself to drive curriculum RL without architectural change.

\paragraph{Compositional video grounding.} Programmatic compositional benchmarks such as CLEVR~\citep{clevr} and GQA~\citep{gqa} generate structured reasoning questions from scene representations, while AGQA~\citep{agqa} is the closest spatio-temporal precedent, generating balanced, difficulty-stratified reasoning questions over Action Genome scene graphs. Two adjacent compositional settings have been explored. \emph{Compositional Temporal Grounding}~\citep{comtg} partitions queries by lexical constituents so that unseen verb-noun-attribute combinations probe language-side generalization on single-clip video. \emph{Compositional Spatio-Temporal Action Localization}, exemplified by Something-Else~\citep{something}, recognizes actions under disjoint train/test verb-noun pairs, while Action Genome~\citep{actiongenome} and Home Action Genome~\citep{home} provide spatio-temporal scene-graph annotations supporting few-shot action recognition. Their target is an action label, not a tube grounded by an arbitrary natural-language description, and neither setting controls multi-dimensional attribute composition against same-category distractors and look-alike time intervals.

\section{Task Formulation}
\label{sec:task}

We formulate Compositional Spatio-Temporal Video Grounding (\textsc{CompSTVG}) as a generalization of standard STVG in which the query is constrained to compose attributes of the target and its relations to other entities without redundancy, so that disambiguation can no longer rely on any single dominant cue. 

\begin{definition}[Compositional STVG]
\label{def:compstvg}
Given an untrimmed video $\mathcal{V}$ and a natural-language query $\mathcal{Q}$, a model predicts a spatio-temporal tube
\[
  \mathcal{T} \;=\; \{\mathcal{T}_k\}_{k=1}^{m}, \qquad
  \mathcal{T}_k \;=\; \bigl\{(t, b_t^{(k)})\bigr\}_{t\in[t_s^{(k)},\,t_e^{(k)}]},
\]
that uniquely localizes the $m\!\geq\!1$ target entities referred to by $\mathcal{Q}$, where each $b_t^{(k)}$ is the bounding box of target $k$ at frame $t$ and $[t_s^{(k)},t_e^{(k)}]$ is its temporal extent. The query is constrained to be a non-redundant composition
\[
  \mathcal{Q} \;=\; \bigl(c_1, c_2, \dots, c_n\bigr), \qquad c_j \in \mathcal{A} \cup \mathcal{R},
\]
of attribute cues $\mathcal{A}$ (e.g., category, appearance, action, temporal pattern) and relation cues $\mathcal{R}$ (e.g., spatial layout and interaction with other entities), such that no proper subset of $\{c_1,\dots,c_n\}$ still uniquely identifies $\mathcal{T}$ on $\mathcal{V}$. 
\end{definition}

In this paper, each $\mathcal{Q}$ is further associated with a difficulty label $d(\mathcal{Q})\in\{1,2,3\}$ that grades disambiguation against the temporal and spatial competitors of $\mathcal{T}$ on $\mathcal{V}$. The non-redundancy constraint guarantees that every cue in $\mathcal{Q}$ is one the model must read to disambiguate, while $d(\mathcal{Q})$ exposes a controllable axis along which performance can be reported instead of a single dataset-level mean. How $d(\mathcal{Q})$ is computed and how queries satisfying both conditions are produced at scale are deferred to Sec.~\ref{sec:bench}, on top of which we build the benchmark \textsc{STVG-CompBench} (Sec.~\ref{sec:bench-eval}) and the trained model \textsc{CurrSTVG} (Sec.~\ref{sec:compstvg}).

\section{Synthesizing Difficulty-Controlled Compositional STVG Data}
\label{sec:bench}

\begin{figure*}[t]
\centering
\includegraphics[width=\linewidth]{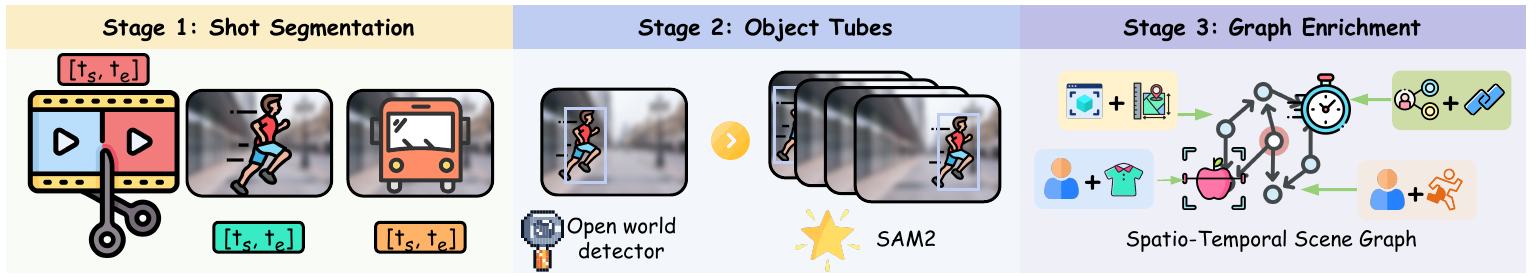}
\caption{The data synthesis pipeline of \textsc{STVG-CompBench}}
\label{fig:pipeline}
\end{figure*}

Realizing Compositional STVG at scale requires data whose queries are simultaneously compositional, minimal, and difficulty-graded. Free-form sentences hand-written by annotators meet none of these constraints in a controllable way, since minimality cannot be enforced post hoc and there is no machinery to grade a query against the competitors next to its target. We therefore build a synthetic data engine on a structured surface, the spatio-temporal scene graph (STSG) of every video. We first define what makes a query hard on the STSG (Sec.~\ref{sec:difficulty}), and then describe how the graph is annotated and how queries of a prescribed difficulty are sampled from it (Sec.~\ref{sec:synthesis}).

\subsection{Spatio-Temporal Scene Graph as Difficulty Measurer}
\label{sec:difficulty}

We adopt a Spatio-Temporal Scene Graph (STSG) as the structured surface on which difficulty is computed:
\begin{equation}
  \mathcal{G} = (V_t,\,V_o,\,V_a,\,E),
  \label{eq:scene-graph}
\end{equation}
where $V_t$ are temporal nodes (intervals $[t_s(v),t_e(v)]$), $V_o$ are object tubes carrying a category $c(o)$, a tube $T_o$, an appearance set $a(o)$, and environment phrases $h(o)$, $V_a$ are action nodes attached to objects with a label $\ell(u)$, target arguments $\mathrm{tar}(u)\subseteq V_o$, and duration $\tau(u)$, and $E$ are relation edges $(o_i,o_j,r_{ij},\mathrm{type}(e),\tau(e))$ whose type is either spatial or non-spatial. Here, $\mathcal{T}$ denotes the complete set of target tubes, and $T_o$ denotes the frame-indexed tube of object $o$.

Random attribute combinations yield mostly redundant queries, while pure graph-structural complexity does not reflect what a grounder must disambiguate. The quantity to measure is the compositional confusion of a query against the competitors next to its target on the graph, namely other intervals of the same object (temporal competitors) and other objects of the same arity (spatial competitors), as defined below.

\begin{definition}[Query Difficulty]
\label{def:difficulty}
A valid query must uniquely point to its target object(s) within the target time interval. For a candidate query $Q=(\mathbf{o}^*,I^*)$ on scene graph $\mathcal{G}$ with target tuple $\mathbf{o}^*$ and target interval $I^*$, let $\mathcal{R}_t(Q)$ denote the set of \emph{temporal competitors} (other intervals of $\mathbf{o}^*$) and $\mathcal{R}_s(Q)$ the set of \emph{spatial competitors} (other candidates with the same arity but different object identities). The query difficulty is defined as
\begin{equation}
  D(Q) \;=\; D_t(Q) + D_s(Q),
  \label{eq:difficulty-total}
\end{equation}
with
\begin{equation}
    D_t(Q) \;=\; 1 - \max_{j\in\mathcal{J}_t(Q)}\;\frac{1}{|\mathcal{R}_t(Q)|}\sum_{r\in\mathcal{R}_t(Q)} E^{t}_{rj},
    \label{eq:difficulty-temporal}
\end{equation}
\begin{equation}
    D_s(Q) \;=\; \frac{X(Q)}{1+X(Q)},\quad
    X(Q) \;=\; \!\!\sum_{r\in\mathcal{R}_s(Q)}\!\! s(Q,r)^{\beta}.
    \label{eq:difficulty-spatial}
\end{equation}
\end{definition}

$D_t(Q)$ is the complement of the best per-attribute exclusion rate against $\mathcal{R}_t(Q)$: the binary indicator $E^{t}_{rj}\in\{0,1\}$ marks whether the temporal attribute $a_j$ rules out competitor $r$, so a higher exclusion rate yields a smaller $D_t$. $D_s(Q)$ measures the analogous confusion against same-arity rivals, where $s(Q,r)$ is a Jaccard--tIoU similarity that grows when $r$ shares more static attributes with $Q$ and overlaps it in time; the exponent $\beta\geq 1$ amplifies a few highly confusable rivals, and the soft-count form keeps $D_s\in[0,1]$ so it is directly comparable with $D_t$. We bin $D(Q)$ into three levels and use the bin index as the difficulty label downstream.

\subsection{From Raw Video to Difficulty-Graded Queries}
\label{sec:synthesis}

With difficulty grounded on the STSG, the data engine factors into two operational stages: annotating $\mathcal{G}$ from a raw video, and sampling, for each prescribed difficulty level, the minimal attribute composition that disambiguates a target.

\subsubsection{STSG Generation}

We orchestrate a set of off-the-shelf specialists in three stages, each of which writes into a distinct slot of $\mathcal{G}$.

\textbf{Stage 1: Shot segmentation.} We cut every video with PySceneDetect~\citep{pyscenedetect} and instantiate each shot as a temporal node in $V_t$ with its start and end frame indices.

\textbf{Stage 2: Object tubes.} Following GroundedSAM2~\citep{grounded}, we couple an open-world detector with SAM2. For each $v_t\in V_t$, the detector runs on the starting frame and feeds boxes to SAM2 for forward propagation across the shot. To capture late-appearing instances, the detector is rerun on clustered keyframes, and every new mask is admitted as a new tube only when its IoU with all existing tracks is low and it does not lie inside any larger mask.

\textbf{Stage 3: Graph enrichment.} The bare tubes are enriched with the attributes downstream queries compose over. For \emph{appearance}, masks are fed to DAM-3B-Video~\citep{dam} for fine-grained captions, which Gemini-3-Flash~\citep{gemini3flash} parses into the category $c(o)$, an appearance attribute set $a(o)$, and an environment phrase $h(o)$. For \emph{actions}, MMAction2~\citep{2mmaction2} tags human tubes with temporal action labels, and Gemini-3-Flash supplements these and tags non-human subjects directly. For \emph{relations}, following Number It~\citep{numberit} and STVG-R1~\citep{stvgr1}, every co-visible pair is overlaid with numeric markers on rendered frames and an MLLM emits a spatial relation and a contact relation. The same visual prompt is reused for cross-shot identity matching to unify tracks into a single $o\in V_o$.

\subsubsection{Difficulty-Controlled Query Sampling}

Since the ability of each attribute and relation cue to rule out a competitor is encoded as a binary indicator $E_{rj}\in\{0,1\}$, cue selection is naturally a 0--1 integer program. For a candidate $Q$ and a query template $f$, we seek the smallest cue subset $x\in\{0,1\}^n$ that excludes every competitor while staying within $f$'s stylistic budget.

The objective minimizes the number of selected cues, so every accepted cue is one the model needs to read:
\begin{equation}
\min_{x\in\{0,1\}^n}\;\sum_{j=1}^{n} x_j.
\label{eq:ilp-obj}
\end{equation}

Each spatial competitor must be ruled out by at least one selected cue, and each temporal competitor by at least $\tau_f$, where $\tau_f$ rises with template difficulty so a single time cue cannot win on its own:
\begin{equation}
\begin{aligned}
\sum_{j} E^{s}_{rj}\,x_j &\geq 1, \quad \forall\,r \in \mathcal{R}_s(Q),\\
\sum_{j} E^{t}_{rj}\,x_j &\geq \tau_f, \quad \forall\,r \in \mathcal{R}_t(Q).
\end{aligned}
\label{eq:ilp-cov}
\end{equation}

Every target-tuple member must be touched by at least one cue, with a non-category cue further required per member when $m\geq 2$ so multi-target queries cannot collapse into a shared class noun:
\begin{equation}
\begin{aligned}
\sum_{j:\,k \in M_j} x_j &\geq 1, \\
\sum_{j:\,k \in M_j,\, c_j \ne c_{\mathrm{cls}}} x_j &\geq \mathbf{1}[m\!\geq\!2],
\end{aligned}
\quad k=1,\dots,m.
\label{eq:ilp-tar}
\end{equation}

Range constraints bound the number of cues of each attribute type $c$ and the overall query length, keeping the template within a recognizable surface form:
\begin{equation}
\begin{aligned}
L_{f,c} \leq \sum_{j:\,c_j=c} x_j &\leq U_{f,c}, \quad \forall\,c \in \mathcal{C},\\
K_f^{\min} \leq \sum_{j} x_j &\leq K_f^{\max}.
\end{aligned}
\label{eq:ilp-style}
\end{equation}

Two scalar constraints push hard templates toward deeper reasoning: $\Gamma_f$ enforces a minimum temporal-chain complexity, and $\rho_f$ requires at least $\rho_f$ cues from a sequentially-qualified subset $\mathcal{S}_f$:
\begin{equation}
\sum_{j} \ell_j\,x_j \geq \Gamma_f,\qquad
\sum_{j \in \mathcal{S}_f} x_j \geq \rho_f.
\label{eq:ilp-depth}
\end{equation}
We collect Eqs.~\eqref{eq:ilp-obj}--\eqref{eq:ilp-depth} as the query sampling program of $(Q,f)$.

We solve this program with \textsc{CompSolver}, a CP-SAT-based constraint solver~\citep{cpsat} tailored to the logical structure of the exclusion constraints, which returns a globally optimal cue subset whenever the $(Q,f)$ pair is feasible. Infeasible pairs are skipped without heuristic fallback. The selected cues are verbalized by an LLM into a single natural-language query, yielding the final $Q$. We predefine 14 query templates spanning single- and multi-target arities. Algorithm~\ref{alg:sampling} summarizes the full sampling loop.

\begin{algorithm}[t]
\caption{Difficulty-Controlled Query Sampling}
\label{alg:sampling}
\begin{algorithmic}[1]
\Require Scene graph $\mathcal{G}$, template pool $\mathcal{F}=\{f_1,\dots,f_{14}\}$, target difficulty bin $d^\star\in\{1,\dots,3\}$
\Ensure Verbalized natural-language query $\mathcal{Q}$ with target tuple $\mathbf{o}^\star$ and interval $I^\star$
\State Sample candidate $(\mathbf{o}^\star,I^\star)$ from $\mathcal{G}$ and a compatible template $f\in\mathcal{F}$
\State Enumerate temporal competitors $\mathcal{R}_t(Q)$ and spatial competitors $\mathcal{R}_s(Q)$ on $\mathcal{G}$
\State Build the cue pool $\{a_j\}_{j=1}^{n}$ and exclusion indicators $E^{t}_{rj},E^{s}_{rj}\in\{0,1\}$
\State Solve the 0--1 ILP in Eqs.~\eqref{eq:ilp-obj}--\eqref{eq:ilp-depth} with \textsc{CompSolver} under the budgets of $f$
\If{infeasible}
    \State \Return \textsc{None} \Comment{skip; no heuristic fallback}
\EndIf
\State Read off the selected cues $x^\star\in\{0,1\}^n$
\State Compute $D(Q)=D_t(Q)+D_s(Q)$ via Eqs.~\eqref{eq:difficulty-temporal}--\eqref{eq:difficulty-spatial}
\If{$\mathrm{bin}(D(Q))\neq d^\star$}
    \State \Return \textsc{None} \Comment{difficulty mismatch}
\EndIf
\State $\mathcal{Q}\leftarrow\textsc{LLM-Verbalize}(\{a_j:x^\star_j=1\},f)$
\State \Return $(\mathcal{Q},\mathbf{o}^\star,I^\star,d^\star)$
\end{algorithmic}
\end{algorithm}

Each accepted run of Algorithm~\ref{alg:sampling} yields one Compositional STVG sample $(\mathcal{V},\mathcal{Q},\mathcal{T},d^\star)$, where $\mathcal{V}$ is the source video, $\mathcal{Q}$ is the verbalized query, the spatio-temporal tube $\mathcal{T}$ is read directly off the target tuple $\mathbf{o}^\star$ and interval $I^\star$ on the STSG, and $d^\star$ is the difficulty bin attached to $\mathcal{Q}$. Iterating the engine over all videos and difficulty bins produces a difficulty-graded corpus that matches the prediction interface of Definition~\ref{def:compstvg}, and we partition it by video into a held-out evaluation split that becomes \textsc{STVG-CompBench} (Sec.~\ref{sec:bench-eval}) and a training split that drives \textsc{CurrSTVG} (Sec.~\ref{sec:compstvg}).

\section{STVG-CompBench: Diagnosing the Compositional Gap}
\label{sec:bench-eval}

\begin{figure*}[t]
\centering
\begin{minipage}{0.245\textwidth}\centering
  \includegraphics[width=\linewidth]{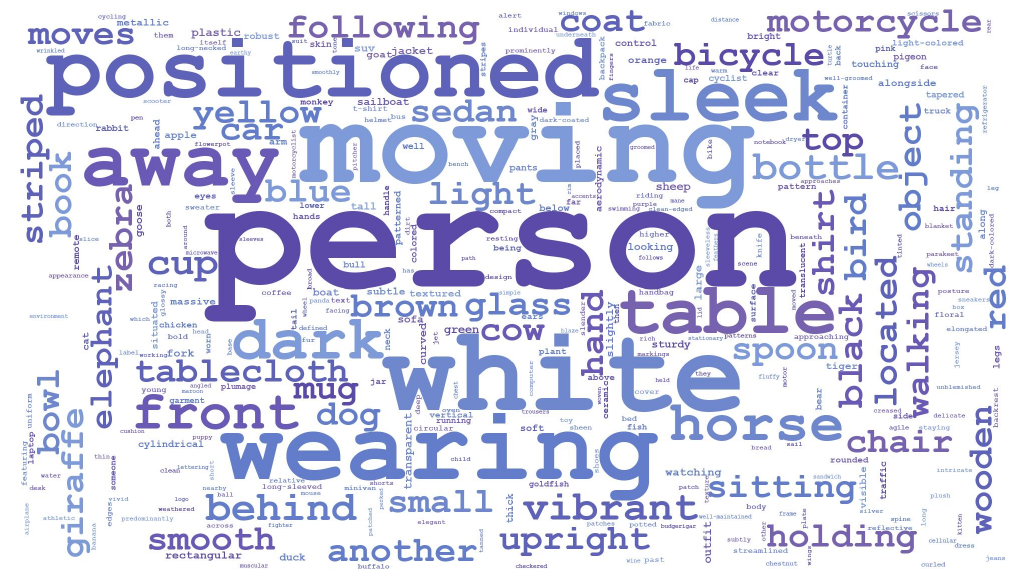}\\
  \footnotesize (a) Query word cloud
\end{minipage}\hfill
\begin{minipage}{0.245\textwidth}\centering
  \includegraphics[width=\linewidth]{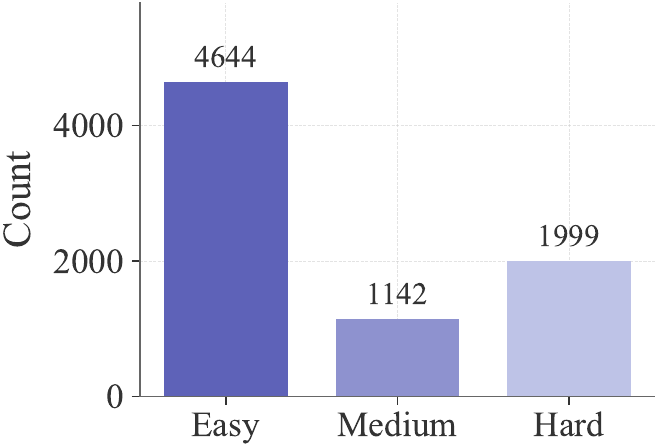}\\
  \footnotesize (b) Difficulty Query Distribution
\end{minipage}\hfill
\begin{minipage}{0.245\textwidth}\centering
  \includegraphics[width=\linewidth]{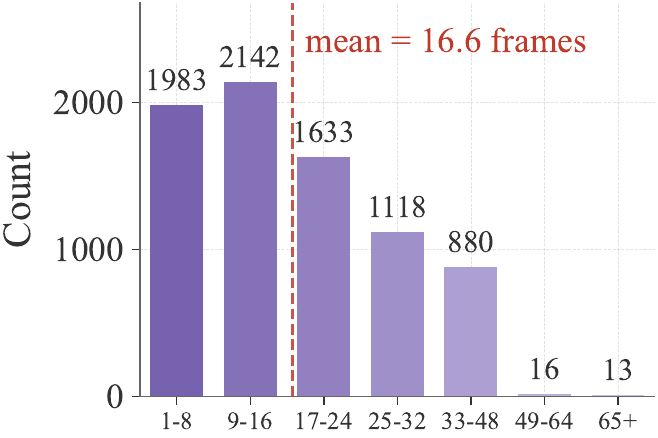}\\
  \footnotesize (c) Grounding Track Length Distribution (frames)
\end{minipage}\hfill
\begin{minipage}{0.245\textwidth}\centering
  \includegraphics[width=\linewidth]{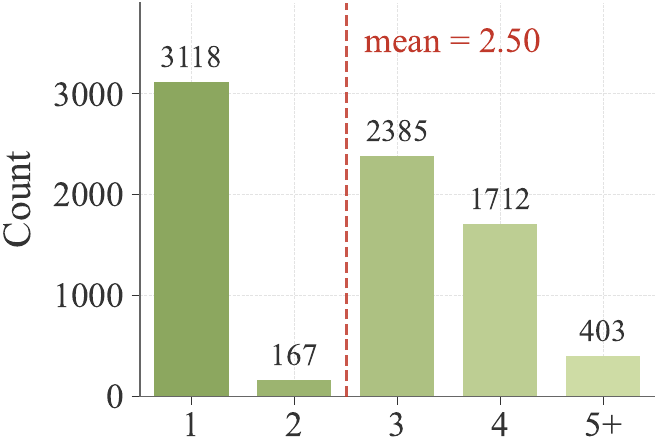}\\
  \footnotesize (d) Attribute Combination Count
\end{minipage}
\caption{Statistics of the whole dataset. (a) Word cloud of query terms. (b) Distribution of queries across the three difficulty buckets of Definition~\ref{def:difficulty}. (c) Distribution of target temporal extent. (d) Distribution of the number of selected cues per query.}
\label{fig:dataset-stats}
\end{figure*}

We organize the empirical study around two questions: (i) does \textsc{STVG-CompBench} expose a compositional gap that the dataset-level average obscures, does grounding accuracy degrade monotonically as the query difficulty $d(\mathcal{Q})$ increases, and (ii) is this degradation consistent across transformer-based and MLLM-based grounder paradigms.

\subsection{Experiment Setup}

\paragraph{Benchmark and Metrics.} \textsc{STVG-CompBench} consists of 2{,}000 high-quality samples drawn from the synthetic corpus, balanced jointly across the three difficulty bins of Definition~\ref{def:difficulty} and across attribute--relation cue combinations, and grouped into three reporting buckets. The full dataset is built on 2{,}000 long, multi-shot source videos sampled from two publicly available corpora, with 1{,}000 videos from MOSE~\citep{mose} and 1{,}000 from the Perception Test~\citep{perceptiontest}. Across these source videos, the engine synthesizes 7{,}785 queries in total: 2{,}000 are selected for \textsc{STVG-CompBench} and the remaining 5{,}785 form the training pool for \textsc{CurrSTVG}. Following standard STVG practice~\citep{vidstg,hcstvg}, we report m\_tIoU, m\_vIoU, vIoU@0.3 and vIoU@0.5 broken down per bucket.

\paragraph{Evaluation Protocol.} All models are run with their officially released checkpoints under a unified 1~FPS frame-sampling protocol; per-frame predictions at any other rate are linearly interpolated to 1~FPS before scoring.

\paragraph{Evaluated Models.}
We benchmark eleven representative grounders covering the two paradigms of Sec.~\ref{sec:related} and our approach \textsc{CurrSTVG}. The vanilla MLLMs are Qwen2.5-VL-7B~\citep{Qwen2.5-VL}, Qwen3-VL-8B~\citep{Qwen3-VL}, Qwen3.5-9B~\citep{qwen3.5}, InternVL3-8B~\citep{internvl3}, InternVL3.5-8B~\citep{internvl3.5} and LLaVA-OneVision-1.5~\citep{LLaVA-OneVision-1.5}, all general-purpose without STVG-specific post-training. The STVG models are the transformer-based grounders TubeDETR~\citep{tubedetr}, CG-STVG~\citep{cgstvg} and TA-STVG~\citep{tastvg}, and the MLLM-based grounders LLaVA-ST~\citep{llavast} and VideoChat-R1~\citep{videochat}.

\subsection{Main Results}

\textsc{STVG-CompBench} exposes a clear performance cliff that the dataset-level average hides (Table~\ref{tab:compbench-main}). Most grounders degrade sharply as difficulty grows: TA-STVG loses 14.2 m\_vIoU from Easy to Hard, TubeDETR drops from 18.01 to 7.00, and several MLLM-based grounders show the same trend, with LLaVA-ST falling from 19.45 to 7.90 and Qwen3.5-9B from 17.32 to 8.16. The high-IoU regime collapses even faster, with vIoU@0.5 dropping from 10--16 on Easy to essentially zero on Hard, so residual m\_vIoU on Hard reflects loose predictions rather than correct tubes. The overall cliff persists across both transformer-based and MLLM-based paradigms, although individual models can deviate from the monotonic trend.

\textsc{CurrSTVG} closes a substantial portion of this gap with the largest absolute gains on the hardest queries. Compared with the strongest baseline LLaVA-ST, \textsc{CurrSTVG} lifts m\_tIoU by 13.3, 28.0 and 28.2 points on Easy, Medium and Hard respectively. The gain grows with difficulty and is largest on Hard, the regime that current STVG benchmarks under-represent, confirming that the difficulty axis of \textsc{STVG-CompBench} is not only diagnostic but actionable.

\begin{table*}[t]
\caption{Performance of representative grounders on \textsc{STVG-CompBench}, broken down by three difficulty buckets of Definition~\ref{def:difficulty} (\textbf{Easy}, \textbf{Medium}, \textbf{Hard}). We report m\_tIoU, m\_vIoU, vIoU@0.3, and vIoU@0.5 (\%). Best results are highlighted in \best{bold}.}
\label{tab:compbench-main}
\centering
\fontsize{7pt}{8.5pt}\selectfont
\setlength{\tabcolsep}{3.2pt}

\begin{adjustbox}{max width=\textwidth}
\begin{tabular}{
  >{\raggedright\arraybackslash}l
  cccc cccc cccc
}
\toprule
\multirow{2}{*}{\textbf{Models}} &
\multicolumn{4}{c}{\textbf{Easy}} &
\multicolumn{4}{c}{\textbf{Medium}} &
\multicolumn{4}{c}{\textbf{Hard}} \\
\cmidrule(lr){2-5}\cmidrule(lr){6-9}\cmidrule(lr){10-13}
& m\_tIoU & m\_vIoU & vIoU@0.3 & vIoU@0.5
& m\_tIoU & m\_vIoU & vIoU@0.3 & vIoU@0.5
& m\_tIoU & m\_vIoU & vIoU@0.3 & vIoU@0.5 \\
\midrule

\rowcolor{dividergray}
\multicolumn{13}{c}{\textit{Open-source Vanilla MLLMs}} \\

Qwen2.5-VL-7B~\citep{Qwen2.5-VL}
  &  6.44 &  0.58 &  0.00 & 0.00  &  3.82 &  0.42 &  0.00 & 0.00  &  3.65 &  0.41 &  0.00 & 0.00 \\
Qwen3-VL-8B~\citep{Qwen3-VL}
  & 58.69 & 12.13 & 14.50 & 3.83  & 33.82 &  6.96 &  3.63 & 0.58  & 28.70 &  6.23 &  2.76 & 0.00 \\
Qwen3.5-9B~\citep{qwen3.5}
  & 63.14 & 17.32 & 21.89 & 9.30  & 39.43 & 11.04 &  9.14 & 1.16  & 30.76 &  8.16 &  5.00 & 0.17 \\
InternVL3-8B~\citep{internvl3}
  & 61.98 &  8.83 &  7.11 & 1.09  & 45.67 &  6.07 &  1.45 & 0.00  & 37.88 &  5.34 &  0.86 & 0.00 \\
InternVL3.5-8B~\citep{internvl3.5}
  & 52.27 &  9.13 & 10.40 & 2.19  & 32.63 &  5.27 &  1.60 & 0.15  & 26.77 &  5.13 &  2.24 & 0.17 \\
LLaVA-OneVision-1.5~\citep{LLaVA-OneVision-1.5}
  & 10.49 &  1.05 &  0.00 & 0.00  &  9.21 &  0.99 &  0.00 & 0.00  & 11.78 &  1.34 &  0.00 & 0.00 \\
\midrule

\rowcolor{dividergray}
\multicolumn{13}{c}{\textit{Open-source STVG Models}} \\

TubeDETR~\citep{tubedetr}
  & 47.82 & 18.01 & 22.57 & 9.71  & 25.41 &  9.04 &  6.24 & 1.31  & 21.31 &  7.00 &  3.79 & 0.00 \\
CG-STVG~\citep{cgstvg}
  & 54.37 &  9.16 &  7.66 & 2.74  & 29.51 &  7.05 &  3.19 & 0.87  & 24.17 &  4.73 &  2.41 & 0.00 \\
TA-STVG~\citep{tastvg}
  & 53.81 & \best{22.91} & 28.32 & \best{16.01} & 28.06 & 11.80 & 10.45 & 1.31  & 22.27 &  8.68 &  6.03 & 0.00 \\
LLaVA-ST~\citep{llavast}
  & 69.16 & 19.45 & 24.21 & 11.90 & 34.23 & 10.14 &  8.13 & 1.31  & 27.08 &  7.90 &  5.69 & 0.00 \\
VideoChat-R1~\citep{videochat}
  & 12.86 &  1.87 &  0.82 & 0.27  &  8.67 &  1.52 &  1.02 & 0.15  &  7.97 &  1.49 &  0.86 & 0.00 \\
\midrule

\rowcolor{ourslavender}
\rowcolor{ourslavender}
\textbf{\textsc{CurrSTVG} (Ours)}
  & \best{82.48} & 20.48 & \best{29.69} & 11.90
  & \best{62.20} & \best{14.85} & \best{16.26} & \best{4.06}
  & \best{55.31} & \best{13.16} & \best{14.14} & \best{2.07} \\
\bottomrule
\end{tabular}
\end{adjustbox}
\end{table*}

\section{CurrSTVG: Curriculum RL on Synthetic Difficulty}
\label{sec:compstvg}

To strengthen MLLM's compositional grounding capability, we turn the difficulty labels of Sec.~\ref{sec:difficulty} into an actionable training signal and propose \textsc{CurrSTVG}, a curriculum reinforcement-learning framework built on Qwen3-VL-8B~\citep{Qwen3-VL}. CurrSTVG is trained in two stages: a short SFT pass internalizing the schema, and a GRPO stage optimizing a composite reward $\mathcal{R}=\mathcal{R}_f+\mathcal{R}_t+\mathcal{R}_s$ that combines a format check, the predicted-interval tIoU, and a temporally-rescaled per-frame box IoU. RL starts on the easiest bucket, and harder buckets are mixed in only after the running reward plateaus.

\subsection{Training strategy of CurrSTVG}

To make the difficulty axis actionable during optimization, we first specify the composite reward used to evaluate each rollout, and then formulate the SFT initialization and GRPO update under an easy-to-hard training schedule.

\paragraph{Reward function.} The composite reward $\mathcal{R}=\mathcal{R}_f+\mathcal{R}_t+\mathcal{R}_s$ has three terms. The format reward $\mathcal{R}_f$ checks that the rollout first emits a textual description of every localized target and then, for each target, a JSON dictionary keyed by frame index whose box coordinates are normalized to $[0,1]$:
\begin{equation}
  \mathcal{R}_{f} = \begin{cases}
  1, & \text{if the output matches the format,} \\
  0, & \text{otherwise.}
  \end{cases}
  \label{eq:compstvg-format}
\end{equation}
The temporal reward takes the first and last frames of the prediction as the predicted interval $\mathcal{T}^p=[t_s,t_e]$, and computes its IoU with the ground truth $\mathcal{T}^{\text{gt}}$,
\begin{equation}
  \mathcal{R}_{t} = \mathrm{IoU}(\mathcal{T}^p,\mathcal{T}^{\text{gt}}).
  \label{eq:compstvg-temporal}
\end{equation}
The spatial reward is computed only on the temporally overlapping frames $\mathcal{F}=\mathcal{T}^p\cap\mathcal{T}^{\text{gt}}$, averaging per-frame IoU on $\mathcal{F}$ and rescaling by the temporal overlap ratio,
\begin{equation}
  \mathcal{R}_{s} = \frac{|\mathcal{F}|}{|\mathcal{T}^p\cup\mathcal{T}^{\text{gt}}|}\cdot\frac{1}{|\mathcal{F}|}\sum_{i\in\mathcal{F}}\mathrm{IoU}(b_i^p,b_i^{\text{gt}}).
  \label{eq:compstvg-spatial}
\end{equation}
The first factor penalizes both extra predicted frames and missed ground-truth frames, while leaving the per-frame term intact when the temporal extent is correct.

\paragraph{SFT stage.} Given a training sample $(\mathcal{V},\mathcal{Q},y)$ with target response $y=(y_1,\dots,y_L)$, the SFT objective is the standard token-level cross-entropy
\begin{equation}
  \mathcal{L}_{\text{SFT}}(\theta) = -\sum_{l=1}^{L}\log \pi_{\theta}\bigl(y_l \,\big|\, \mathcal{V},\mathcal{Q},y_{<l}\bigr).
  \label{eq:compstvg-sft}
\end{equation}

\paragraph{GRPO stage.} Starting from the SFT checkpoint $\pi_{\theta_0}$, we sample $N=8$ responses $\{o_i\}_{i=1}^{N}$ per training sample, each scored by $\mathcal{R}(o_i)$. The advantage is computed by group-wise normalization,
\begin{equation}
  A_i = \frac{\mathcal{R}(o_i) - \mathrm{mean}(\{\mathcal{R}(o_j)\}_{j=1}^{N})}{\mathrm{std}(\{\mathcal{R}(o_j)\}_{j=1}^{N})},
\end{equation}
and the policy is updated against the frozen reference $\pi_{\theta_0}$ with a clipped importance ratio and KL regularization,
\begin{equation}
\begin{aligned}
\mathcal{J}_{\text{GRPO}}(\theta) = \mathbb{E}\Big[\tfrac{1}{N}\!\sum_{i=1}^{N}\!\big(&\min(\rho_i A_i, \\
&\mathrm{clip}(\rho_i,1\!-\!\epsilon,1\!+\!\epsilon)A_i)\\
&-\,\beta\,\mathrm{KL}(\pi_\theta\|\pi_{\text{ref}})\big)\Big],
\end{aligned}
\end{equation}
where $\rho_i(\theta)=\pi_\theta(o_i\!\mid\!\mathcal{V},\mathcal{Q})/\pi_{\theta_{\text{old}}}(o_i\!\mid\!\mathcal{V},\mathcal{Q})$.

\subsection{Experiment Setup}

\paragraph{Training Configuration.}
We instantiate \textsc{CurrSTVG} on Qwen3-VL-8B over the training pool of Sec.~\ref{sec:synthesis}, balanced across difficulty buckets and split into SFT and RL subsets at $5\!:\!2$. GRPO runs on 8 A100 GPUs with $N=8$ rollouts per sample under the same 1~FPS sampling protocol as Sec.~\ref{sec:bench-eval}, advancing a difficulty bucket only after the running reward plateaus.

\paragraph{Baselines.} We compare \textsc{CurrSTVG} against representative MLLM-based grounders, including GroundingGPT~\citep{groundinggpt}, Qwen2.5-VL~\citep{Qwen2.5-VL}, GPT-4o~\citep{gpt4o}, LLaVA-ST~\citep{llavast}. On \textsc{STVG-CompBench} we additionally retain the eleven grounders of Sec.~\ref{sec:bench-eval} so that the in-domain comparison covers both transformer-based and MLLM-based paradigms.

\paragraph{Benchmarks and Evaluation.}
We evaluate under two settings. The in domain setting reports on \textsc{STVG-CompBench}, where queries follow the same distribution as the training pool but no video is shared between splits. The out-of-domain setting reports on two existing STVG benchmarks whose training data \textsc{CurrSTVG} never sees. \textbf{HC-STVG-v2}~\citep{hcstvg} focuses on multi-person scenes, has 10,131 training, 2,000 validation, and 4,413 test samples.
Since the test annotations for v2 are not public, we report results on the validation set, following prior work~\citep{tastvg}; and \textbf{VidSTG}~\citep{vidstg}, which contains 99{,}943 declarative and interrogative sentences over 80 object categories on 6{,}924 untrimmed videos and on which we report the 10{,}053-query test set under the standard split~\citep{vidstg}, with declarative and interrogative subsets reported separately. All three benchmarks reuse the metric suite of Sec.~\ref{sec:bench-eval} under the same 1~FPS frame sampling protocol.

\subsection{Generalization to Established STVG Benchmarks}

\textsc{CurrSTVG} transfers to standard STVG benchmarks despite never seeing their training data. On HC-STVG-v2 (Table~\ref{tab:hcstvg}), it reaches the best m\_tIoU at 46.7 and improves the high-IoU regime over its Qwen3-VL backbone by roughly an order of magnitude (vIoU@0.3: 6.2$\to$12.3, vIoU@0.5: 0.2$\to$2.32). On VidSTG (Table~\ref{tab:vidstg}), it remains competitive on most metrics; interrogative sentences generally trail declarative ones on m\_vIoU and vIoU, although temporal-only scores can differ.

\begin{table}[!htbp]
\centering
\scriptsize
\setlength{\tabcolsep}{3pt}
\caption{Comparison on HC-STVG-v2 (\%). Best in \best{bold}.}
\label{tab:hcstvg}
\begin{adjustbox}{max width=\linewidth}
\begin{tabular}{lcccc}
\toprule
\textbf{Model} & \textbf{m\_tIoU} & \textbf{m\_vIoU} & \textbf{vIoU@0.3} & \textbf{vIoU@0.5} \\
\midrule
GPT-4o~\citep{gpt4o}                   & 32.7 &  9.1 &  5.7 &  0.0 \\
GroundingGPT~\citep{groundinggpt} & 19.6 & \best{14.7} & \best{16.6} & 3.1 \\
Qwen2.5-VL-7B~\citep{Qwen2.5-VL}  & 22.9 & 13.0 & 15.6 & \best{6.4} \\
Qwen3-VL-8B~\citep{Qwen3-VL}      & 43.2 & 11.2 &  6.2 &  0.2 \\
\midrule
\rowcolor{ourslavender}
\textbf{\textsc{CurrSTVG} (Ours)} & \best{46.7} & 13.1 & 12.3 & 2.32 \\
\bottomrule
\end{tabular}
\end{adjustbox}
\end{table}

\begin{table}[!htbp]
\centering
\scriptsize
\setlength{\tabcolsep}{2.5pt}
\caption{Comparison on VidSTG (\%) for declarative and interrogative sentences. Best in \best{bold}.}
\label{tab:vidstg}
\begin{adjustbox}{max width=\linewidth}
\begin{tabular}{lcccccccc}
\toprule
\multirow{2}{*}{\textbf{Model}} &
\multicolumn{4}{c}{\textbf{Declarative}} &
\multicolumn{4}{c}{\textbf{Interrogative}} \\
\cmidrule(lr){2-5}\cmidrule(lr){6-9}
& m\_tIoU & m\_vIoU & v@.3 & v@.5
& m\_tIoU & m\_vIoU & v@.3 & v@.5 \\
\midrule
GPT-4o~\citep{gpt4o}            & \best{38.3} &  9.2 &  7.1 &  1.6 & \best{39.8} &  6.1 &  3.5 &  0.6 \\
GroundingGPT~\citep{groundinggpt} & 15.5 & \best{12.3} & 13.2 & 4.1 & 11.9 &  \best{8.7} &  9.6 &  2.9 \\
Qwen2.5-VL~\citep{Qwen2.5-VL}   & 16.8 & 10.9 & \best{14.3} & \best{5.4} & 13.8 &  8.5 & \best{11.3} &  \best{4.4} \\
\midrule
\rowcolor{ourslavender}
\textbf{\textsc{CurrSTVG} (Ours)} & 31.41 &  9.15 & 10.29 &  5.05 & 30.88 &  6.12 &  5.98 &  2.75 \\
\bottomrule
\end{tabular}
\end{adjustbox}
\end{table}

\subsection{Ablation on Curriculum and Difficulty Labels}
\label{sec:abl-curriculum}

To verify that the gains of \textsc{CurrSTVG} come from the difficulty axis rather than extra synthetic data, we compare easy-to-hard curriculum training against shuffled training on the same pool. The curriculum variant clearly outperforms shuffled training on \textsc{STVG-CompBench}, and the gap concentrates on the hardest bucket. Under shuffled training, hard queries are encountered before the policy has stabilized on the format and temporal rewards, so the spatial reward on dense compositions provides a noisy gradient that the policy never fully exploits; the curriculum lets these rewards stabilize on simpler queries first and reserves capacity for the compositions that drive the cliff observed on \textsc{STVG-CompBench}.

\subsection{Human Verification of STSG Construction}
\label{sec:human-verification}

We validate the reliability of our STSG construction pipeline through human verification. We randomly sample 300 queries, stratified by template and difficulty, and manually verify them against the full original videos, judging every query item by item along the dimensions corresponding to the STSG stages. The resulting accuracies are reported in Table~\ref{tab:human-verification}.

\begin{table}[!htbp]
\centering
\footnotesize
\setlength{\tabcolsep}{4pt}
\caption{Human verification of the STSG construction pipeline on 300 randomly sampled queries.}
\label{tab:human-verification}
\begin{adjustbox}{max width=\linewidth}
\begin{tabular}{@{}>{\raggedright\arraybackslash}p{0.39\linewidth} >{\raggedright\arraybackslash}p{0.24\linewidth} r r@{}}
\toprule
\textbf{STSG stage} & \textbf{Verified dimension} & \textbf{Items} & \textbf{Accuracy} \\
\midrule
Detection + SAM2 tracking & Object trajectory & 478 & 94.7\% \\
Caption / action parsing & Attribute & 410 & 91.7\% \\
Relation extraction & Relation & 377 & 93.6\% \\
Query uniqueness (incl. cross-shot identity) & Unique reference & 478 & 85.4\% \\
Temporal grounding & Temporal range & 478 & 80.3\% \\
\bottomrule
\end{tabular}
\end{adjustbox}
\end{table}

Human annotators report 94.7\% accuracy for object trajectories, 91.7\% for attributes, and 93.6\% for relations. For the two uniqueness-critical stages, they report 85.4\% accuracy for unique reference and 80.3\% for temporal range. Among all errors, about 60\% stem from objective difficulty such as occlusion and ambiguous temporal boundaries, and only 32\% are fabricated or mismatched content, which amounts to merely 3.1\% of all judged items. These results indicate that our STSG construction pipeline is sufficiently reliable and that the resulting annotations are highly faithful to the original videos.

\FloatBarrier

\subsection{Ablation on the Constraint Programming Solver}
\label{sec:solver-ablation}

The solver in the data engine enforces non-redundancy on every synthesized query; a weaker solver translates directly into redundant cues. We compare \textsc{CompSolver} against three alternatives on the same binary integer programming problem: a generic ILP backend~\citep{wolsey2020} using a standard mixed-integer solver, a set-cover greedy heuristic~\citep{chvatal1979} that adds the cue excluding the most competitors until coverage in Eq.~\eqref{eq:ilp-cov} is met, and a random sampler that accepts the first feasible cue subset under the size budget of $f$. All four are run on 338{,}168 candidate query--template pairs, reporting success rate, optimality rate, objective gap and ratio against the optimum, worst-case ratio, and mean / 95th-percentile latency~\citep{wolsey2020}.

\begin{table}[!htbp]
\centering
\small
\caption{Comparison of constraint programming solvers on 338{,}168 candidate query–template pairs.  Best results are in \best{bold}.}
\label{tab:solver-ablation}
\setlength{\tabcolsep}{4.0pt}
\begin{adjustbox}{max width=\linewidth}
\begin{tabular}{lcccccrr}
\toprule
\textbf{Solver} & \textbf{Succ\%} & \textbf{Opt\%} & \textbf{Gap\%} & \textbf{Ratio} & \textbf{R\textsubscript{max}} & \textbf{$t_{\text{mean}}$} & \textbf{$t_{p95}$} \\
\midrule
Random  & 27.2 & 24.6 & 39.9 & 1.40 & 3.00 &  19.7 &  53.2 \\
Greedy  & 50.5 & 41.1 & 26.9 & 1.27 & 2.00 &   \best{0.7} &   \best{1.9} \\
ILP     & \best{66.2} & \best{100.0} & \best{0.0} & \best{1.00} & \best{1.00} &  81.2 & 327.1 \\
\midrule
\rowcolor{ourslavender}
\textsc{CompSolver} (Ours) & \best{66.2} & \best{100.0} & \best{0.0} & \best{1.00} & \best{1.00} & 27.5 & 100.8 \\
\bottomrule
\end{tabular}
\end{adjustbox}
\end{table}

Table~\ref{tab:solver-ablation} shows that exact solvers are essential and \textsc{CompSolver} delivers the best speed–quality trade-off. Random and greedy succeed on only 27.2\% and 50.5\% of feasible pairs and incur 39.9\% and 26.9\% optimality gaps, both heuristics violate the minimality requirement on a non-trivial fraction of queries and let redundant cues into the data. The generic ILP backend recovers the optimum on every feasible pair but pays 81.2 ms mean and 327.1 ms tail latency. \textsc{CompSolver} matches the ILP backend on every quality metric while running roughly three times faster.

\FloatBarrier
\section{Conclusion}

We propose Compositional STVG, a task that requires every query to compose attribute and relation cues without redundancy and to be graded by an explicit difficulty label. We further propose a synthetic data engine that casts non-redundant cue selection as a constraint programming problem over a spatio-temporal scene graph, and use it to build the diagnostic benchmark \textsc{STVG-CompBench} and the curriculum-RL model \textsc{CurrSTVG}. By tying the same difficulty axis to both benchmark construction and training, the framework makes compositionality measurable, controllable, and actionable rather than treating it as an unobserved property of the data. \textsc{CurrSTVG} delivers the largest gains exactly where the compositional cliff is steepest, while the remaining transfer and short-track limitations motivate future work on broader data coverage and more robust temporal reasoning.

\section*{Limitations}

Three scope boundaries remain. First, out-of-domain generalization is imperfect. \textsc{CurrSTVG} transfers to HC-STVG-v2 and VidSTG without seeing their training data but still trails the strongest specialist baselines. Second, performance degrades on very short target tracks, where the temporally-rescaled spatial reward concentrates on too few frames to give a stable gradient. Third, the IoU-based reward shape can incentivize over-long outputs, since widening the predicted interval and padding the JSON reference are not penalized by the format check; we partially mitigate this with rollout-length penalties.

\bibliography{custom}

\appendix

\clearpage
\section*{Appendix}

\section{Evaluated Models}
\label{app:models}

We benchmark a wide set of grounders on \textsc{STVG-CompBench}, covering general-purpose vision--language MLLMs as well as specialist STVG models. All models are run with their officially released checkpoints under the unified protocol of Sec.~\ref{sec:bench-eval}; we adapt only their input schema so that each model receives the same video frames and natural-language query, and emits a spatio-temporal tube in its native format. We group the evaluated models into the following two families.

\paragraph{Open-source vanilla MLLMs.} These are general-purpose vision--language models without STVG-specific post-training. They are prompted to emit per-frame bounding boxes and a temporal interval in a JSON schema; predictions are post-processed into a tube without any additional training.
\begin{itemize}\setlength{\itemsep}{1pt}
  \item \textbf{Qwen2.5-VL-7B}~\citep{Qwen2.5-VL} -- a multilingual vision--language model with native bounding-box output, included as a vanilla MLLM baseline.
  \item \textbf{Qwen3-VL-8B}~\citep{Qwen3-VL} -- the next-generation Qwen vision--language model used both as a baseline and as the backbone of \textsc{CurrSTVG}.
  \item \textbf{Qwen3.5-9B}~\citep{qwen3.5} -- a stronger Qwen3.5 variant with extended visual reasoning, evaluated under the same prompting protocol.
  \item \textbf{InternVL3-8B}~\citep{internvl3} -- a vision--language model trained with progressive multi-task alignment, included to represent a non-Qwen MLLM family.
  \item \textbf{InternVL3.5-8B}~\citep{internvl3.5} -- the InternVL3.5 release that extends InternVL3 with stronger visual reasoning.
  \item \textbf{LLaVA-OneVision-1.5}~\citep{LLaVA-OneVision-1.5} -- the OneVision branch of LLaVA targeted at unified single- and multi-image understanding; included to test a non-grid frame protocol.
\end{itemize}

\paragraph{Open-source STVG models.} These are models explicitly trained or fine-tuned on spatio-temporal video grounding, ranging from transformer-based one-stage grounders to MLLM-based grounders.
\begin{itemize}\setlength{\itemsep}{1pt}
  \item \textbf{TubeDETR}~\citep{tubedetr} -- a transformer-based one-stage grounder that regresses the full spatio-temporal tube end-to-end.
  \item \textbf{CG-STVG}~\citep{cgstvg} -- a context-guided one-stage grounder that injects cross-modal context into a DETR-style decoder.
  \item \textbf{TA-STVG}~\citep{tastvg} -- a target-aware grounder that uses target-specific queries to disambiguate distractors during decoding.
  \item \textbf{LLaVA-ST}~\citep{llavast} -- a spatio-temporal extension of LLaVA that injects learnable spatio-temporal tokens into the VLM and supervises both interval and box prediction.
  \item \textbf{VideoChat-R1}~\citep{videochat} -- an MLLM grounder fine-tuned with multi-task GRPO under an IoU reward.
\end{itemize}
For models whose evaluation cells appear blank in Table~\ref{tab:compbench-main} (e.g.\ LLaVA-NeXT-Video, LLaVA-OneVision-2, GroundingGPT, DeViL, VTimeLLM, Grounded-VideoLLM), evaluation is in progress at the time of submission and is not used to support any quantitative claim.

\section{Evaluation Metrics}
\label{app:metrics}

We follow the standard STVG evaluation protocol~\citep{vidstg,hcstvg} and report four metrics, organized as one temporal metric and three spatio-temporal metrics. Let $\mathcal{T}^p=[t_s^p,t_e^p]$ and $\mathcal{T}^{\text{gt}}=[t_s^{\text{gt}},t_e^{\text{gt}}]$ denote the predicted and ground-truth time intervals respectively, and $b_i^p,b_i^{\text{gt}}$ the predicted and ground-truth boxes on frame $i$.

\paragraph{m\_tIoU.} The mean temporal Intersection-over-Union, averaged over all queries:
\begin{equation}
  \mathrm{m\_tIoU} = \frac{1}{|\mathcal{D}|}\sum_{q\in\mathcal{D}}\frac{|\mathcal{T}^p_q\cap\mathcal{T}^{\text{gt}}_q|}{|\mathcal{T}^p_q\cup\mathcal{T}^{\text{gt}}_q|}.
\end{equation}

\paragraph{m\_vIoU.} The mean video Intersection-over-Union, which composes temporal alignment with per-frame box overlap by averaging frame-wise IoU only on temporally overlapping frames and rescaling by the temporal-union ratio:
\begin{equation}
  \mathrm{vIoU}_q = \frac{1}{|\mathcal{T}^p_q\cup\mathcal{T}^{\text{gt}}_q|}\sum_{i\in\mathcal{T}^p_q\cap\mathcal{T}^{\text{gt}}_q}\mathrm{IoU}(b_i^p,b_i^{\text{gt}}),
\end{equation}
and $\mathrm{m\_vIoU}=|\mathcal{D}|^{-1}\sum_q \mathrm{vIoU}_q$.

\paragraph{vIoU@$\tau$.} The recall at vIoU threshold $\tau$, defined as the fraction of queries whose vIoU exceeds $\tau$:
\begin{equation}
  \mathrm{vIoU@}\tau = \frac{1}{|\mathcal{D}|}\sum_{q\in\mathcal{D}}\mathbf{1}[\mathrm{vIoU}_q\geq\tau].
\end{equation}
We report $\tau\in\{0.3,0.5\}$, where vIoU@0.5 captures the high-precision regime that matters for downstream tube consumption (e.g.\ video editing and embodied perception).

All four metrics are reported both at the dataset level and broken down per difficulty bucket of Definition~\ref{def:difficulty}, so that a model whose dataset-level number is dominated by a long tail of single-cue queries cannot mask its compositional weakness.

\section{Benchmark Details}
\label{app:bench}

\paragraph{Source videos.} The full dataset is built on 2{,}000 long, multi-shot source videos sampled from two publicly available corpora, with 1{,}000 videos from MOSE~\citep{mose} and 1{,}000 from the Perception Test~\citep{perceptiontest}. Each video is processed by the data engine of Sec.~\ref{sec:synthesis} to yield its spatio-temporal scene graph, on top of which queries are sampled by the ILP-based procedure of Algorithm~\ref{alg:sampling}.

\paragraph{Splits.} The 7{,}785-query corpus is split into a 2{,}000-query evaluation set and a 5{,}785-query training pool with disjoint videos. The evaluation set is balanced across the three difficulty levels of Definition~\ref{def:difficulty}, and the training pool is also balanced across the three levels by construction. No video, scene graph, or query is shared between the two splits.

\paragraph{Inference protocol.} For all evaluated models, video frames are sampled at 1~FPS, capped at the maximum frame count supported by each model's released configuration. Models that natively predict per-frame boxes emit them at 1~FPS; models that predict at a different frame rate are linearly interpolated to 1~FPS before scoring. The same protocol is used at training time for \textsc{CurrSTVG}, so training and evaluation operate at a matched temporal resolution.

\paragraph{External benchmarks.} For the generalization study in Sec.~\ref{sec:compstvg}, we additionally evaluate on two standard STVG benchmarks. \textbf{HC-STVG-v2}~\citep{hcstvg} focuses on multi-person scenes where each untrimmed video is paired with a textual description of human attributes and actions, and contains 10{,}131 / 2{,}000 / 4{,}413 train/val/test samples. Following prior work~\citep{cgstvg,tastvg}, we report on the validation set since v2 test annotations are not public, and use the same split. After removing corrupted videos in our environment, evaluation is run on 1{,}895 videos and queries. \textbf{VidSTG}~\citep{vidstg} consists of 6{,}924 untrimmed videos with 99{,}943 declarative and interrogative sentences referring to 80 object categories, forming 44{,}808 video–triplet instances; following the standard split~\citep{vidstg}, it uses 5{,}563 / 618 / 743 videos and 80{,}684 / 8{,}956 / 10{,}303 sentences for train/val/test. After removing corrupted videos, our test set contains 10{,}053 queries with declarative and interrogative subsets reported separately. Both benchmarks reuse the metric suite of App.~\ref{app:metrics}, evaluated under the same 1~FPS frame-sampling protocol used elsewhere in this paper.

\section{Prompt}
\label{app:prompt}

We list the prompt templates used by the data engine of Sec.~\ref{sec:synthesis} for STSG annotation: structured attribute extraction from per-object captions, spatial relation extraction over object pairs, non-spatial (temporal) relation extraction, and cross-shot identity matching for tube unification.

\begin{promptbox}{Structured Attribute Extraction Prompt}
\textbf{System Prompt:} You are a precise information extraction assistant.

\textbf{User Prompt:}
You are an expert in scene understanding. I will give you a short paragraph that describes a video clip.

Your task is to extract structured information about a single object described in the paragraph. Be careful not to omit any representative object information.

Please return a JSON with the following fields:
\begin{itemize}\setlength{\itemsep}{0pt}
  \item \texttt{"object"}: The main object being described (e.g., \texttt{"person"}, \texttt{"dog"}, \texttt{"car"}). If the inference about the object is uncertain based on the description, add \texttt{"(uncertain)"} after the object name.
  \item \texttt{"attributes"}: A list of ONLY the visual/physical attributes that can be directly observed about the object itself. Include only:
  \begin{itemize}\setlength{\itemsep}{0pt}
    \item Visual appearance: color, shape, size, texture, pattern, material appearance, style, the clothing and appearance of the person.
    \item Physical properties: state, transparency, reflectiveness, orientation, material.
    \item Design elements: stripes, dots, logos, decorative features.
  \end{itemize}
  DO NOT include implied states, inferred conditions, functional descriptions, or anything that describes the object's interaction with its environment.
  \item \texttt{"environment"}: A list of environment/context relations between this object and other entities (but not clothes and actions), e.g., \texttt{"on top of table"}, \texttt{"next to person"}, \texttt{"inside container"}, \texttt{"facing camera"}, \texttt{"part of group"}, \texttt{"leaning against the wall"}, \texttt{"carrying a briefcase"}.
  \item \texttt{"actions"}: A list of actions that the object is performing or movements it is making (e.g., \texttt{"rotating"}, \texttt{"moving"}, \texttt{"falling"}, \texttt{"bouncing"}, \texttt{"sliding"}, \texttt{"right arm extended outward"}). Including the subtle movements of the person.
\end{itemize}

Important distinctions:
\begin{itemize}\setlength{\itemsep}{0pt}
  \item Attributes = What the object looks like (visual only). Please use ADJECTIVE form. If you are extracting a person's clothing, put it in attributes.
  \item Environment = How the object relates to other things spatially, functionally, or contextually. If you are extracting interactions between objects and other objects or the environment, put them in environment.
  \item Actions = What the object is doing or how it's moving.
  \item However, please note that the attributes of clothing on a person should not be directly stored as attributes of the person. If the description mentions a brown hat, it should be stored as ``wear a brown hat'' rather than just ``brown''.
\end{itemize}

Now process the following description:\\
\texttt{\{description\}}
\end{promptbox}

\begin{promptbox}{Spatial Relation Extraction Prompt}
Role\\
You are a detail-oriented Video Relationship Annotator responsible for reviewing sequences of sampled video frames and extracting a comprehensive set of spatial relationships. All relationships must be visually grounded, type-consistent, and strictly follow the defined schema.

Task Context
\begin{itemize}\setlength{\itemsep}{0pt}
  \item Videos are sampled at 1 fps.
  \item Each frame already has the two target objects marked by red integer IDs.
  \item The red number at the lower-right corner of each frame is the frame index. You must use these frame indices to determine the time spans of relationships.
  \item You must use the marked IDs directly and must not detect new objects.
  \item Your analysis should consider all provided frames jointly.
\end{itemize}

Input Format
\begin{itemize}\setlength{\itemsep}{0pt}
  \item Input is an ordered sequence of frames from one video.
  \item In each frame, exactly two target objects are marked: Object A (\texttt{id=\{id\_a\}}, \texttt{class=\{class\_a\}}) and Object B (\texttt{id=\{id\_b\}}, \texttt{class=\{class\_b\}}).
  \item Frame indices in this sequence: \texttt{\{frame\_ids\}}.
\end{itemize}

Guidelines
\begin{itemize}\setlength{\itemsep}{0pt}
  \item Extract only purely spatial (physical or geometric) relationships visible in the 3D scene. Do not extract any relationships that indicate state, function, action, or purpose.
  \item Exclude all temporal, social, functional, or attentional relationships, as well as any stateful or action-based verbs.
  \item Each relationship must be visually supported by the visual frames.
  \item Ensure logical consistency with common sense and real-world physics; do NOT output implausible or unsupported relationships.
  \item Think in terms of 3D spatial layout by using depth information derived from world knowledge and visual cues, not just 2D image positions. Do NOT rely solely on 2D bounding box coordinates. Do NOT output \texttt{"left of"} or \texttt{"right of"}.
  \item Use precise, explicit, and non-redundant verbs.
  \item Do NOT miss any clear and valid spatial relationships between objects.
\end{itemize}

Temporal Grounding\\
Output relationships with one or more time spans \texttt{[[start\_frame, end\_frame], ...]}, as continuous intervals where the relationship is visually supported and both objects are present.

Output Format\\
You must only return a single valid JSON object strictly following this schema:
\begin{verbatim}
{
  "relationships": [
    [subject_id, predicate_verb, object_id,
     [[start_frame, end_frame], ...]]
  ]
}
\end{verbatim}

Output Specifications
\begin{itemize}\setlength{\itemsep}{0pt}
  \item \texttt{subject\_id}/\texttt{object\_id} must be integers and should be either \texttt{\{id\_a\}} or \texttt{\{id\_b\}}.
  \item If there are no valid relationships, output: \texttt{\{"relationships": []\}}.
  \item Output strict valid JSON only, with key \texttt{"relationships"}.
\end{itemize}
\end{promptbox}

\begin{promptbox}{Non-Spatial Relation Extraction Prompt}
Role\\
You are a detail-oriented \textbf{Video Relationship Annotator} tasked with reviewing sequences of sampled video frames and extracting a comprehensive set of \textbf{temporal (non-spatial) relationships}. Ensure all extracted relationships are visually grounded, type-consistent, and strictly follow the defined schema. You are analyzing videos sampled at 1 fps; each frame contains detected objects with bounding boxes. Analyze all frames jointly and output temporal relationships.

Task Context
\begin{itemize}\setlength{\itemsep}{0pt}
  \item Videos are sampled at 1 fps.
  \item Each frame already has the two target objects marked by red integer IDs.
  \item The red number at the lower-right corner of each frame is the frame index.
  \item Two target objects are already marked in red: Object A (\texttt{id=\{id\_a\}}, \texttt{class=\{class\_a\}}) and Object B (\texttt{id=\{id\_b\}}, \texttt{class=\{class\_b\}}).
  \item Frame indices in this sequence: \texttt{\{frame\_ids\}}.
\end{itemize}

Relationship Taxonomy: Classify each relationship into exactly ONE category:
\begin{enumerate}\setlength{\itemsep}{0pt}
  \item \textbf{Functional --- Contact / Manipulation}: direct physical interaction where an animate subject alters or uses the state of another object. Subject: animate; Object: animate or inanimate. Exclude pure motion or gaze without contact.
  \item \textbf{Stateful --- Attachment / Possession-like}: visually grounded, time-persistent attachment or carrying relationships indicating sustained physical association rather than instantaneous action. Exclude abstract ownership or purely spatial layout.
  \item \textbf{Motion --- Relative Movement}: temporal changes in relative position or movement trajectory between entities. Subject: movable; Object: animate or inanimate. Exclude static layout or manipulation actions.
  \item \textbf{Social --- Animate-to-Animate Interaction}: communication, coordination, or interpersonal acts between animate agents. Both subject and object must be animate. Exclude one-sided attention or non-social contact.
  \item \textbf{Attentional --- Gaze / Focus (includes Camera)}: visual attention or camera focus directed at another object or agent. Subject: animate or camera (with \texttt{object\_id = -1}); Object: animate or inanimate. Exclude communication or manipulation.
  \item \textbf{Event-Level --- Goal-Directed Multi-Step Activity}: higher-level, time-extended actions combining multiple functional, causal or motion relations into a single purposeful event. Exclude single short actions or ungrounded intent.
\end{enumerate}

Core Analysis Logic \& Constraints
\begin{enumerate}\setlength{\itemsep}{0pt}
  \item \textbf{Object typing:} Animate (humans, animals, humanoid robots); Inanimate (cars, tools, furniture, etc.); Camera (unseen observer/recorder, always \texttt{object\_id = -1}).
  \item \textbf{Typing rules:} Functional/Social subject must be animate; Motion subject must be movable; Attentional subject must be animate; Social requires both subject and object animate.
  \item \textbf{Extraction basis:} all relationships must be visually supported and logically consistent with common sense; do NOT infer relationships not visually evidenced. Relationships must have temporal grounding (one or more continuous frame intervals supported by visual evidence). End the relationship at the object's last visible frame; do not continue under occlusion. Do NOT create self-relations (\texttt{subject\_id == object\_id}).
\end{enumerate}

Output Format\\
Return one valid JSON object:
\begin{verbatim}
{
  "relationships": [
    [subject_id, predicate_verb, object_id,
     [[start_frame, end_frame], ...],
     relationship_type]
  ]
}
\end{verbatim}

Output Details
\begin{itemize}\setlength{\itemsep}{0pt}
  \item \texttt{relationship\_type} must be one of: \texttt{functional}, \texttt{stateful}, \texttt{motion}, \texttt{social}, \texttt{attentional}, \texttt{event\_level}.
  \item If no valid relation exists, output \texttt{\{"relationships": []\}}.
  \item Output strict valid JSON only, with key \texttt{"relationships"}.
\end{itemize}
\end{promptbox}

\begin{promptbox}{Cross-Shot Identity Matching Prompt}
\textbf{System Prompt:} You are a rigorous visual verifier. Carefully inspect all provided paired images, cross-check consistency before deciding, and output strict JSON only.

\textbf{User Prompt:}\\
Role\\
You are a careful \textbf{cross-shot entity matching annotator}. Your goal is to match ids in Shot A to ids in Shot B only when visual evidence is strong.

Task Context
\begin{itemize}\setlength{\itemsep}{0pt}
  \item You receive 3 paired images from the same video.
  \item In each paired image: left = Shot A, right = Shot B.
  \item Candidate objects are marked with red integer ids.
  \item Id numbers are local within each shot and may differ across shots.
  \item The same physical object should keep consistent appearance cues across the 3 pairs.
  \item Shot A candidate ids: \texttt{\{shot\_a\_ids\}}. Shot B candidate ids: \texttt{\{shot\_b\_ids\}}.
\end{itemize}

Visibility Notes\\
\texttt{\{visibility\_notes\}}

Required Analysis Checklist
\begin{itemize}\setlength{\itemsep}{0pt}
  \item Compare each proposed match across all pairs, not just one image.
  \item Check stable cues: body shape, clothing/texture, size, accessories, relative position/motion pattern.
  \item Reject pairs with weak or ambiguous evidence.
  \item The input image pairs are sorted in time order, so there may not be matching objects in some pairs. Reason with temporal relationships before deciding; do not force matches.
\end{itemize}

Output Rules
\begin{itemize}\setlength{\itemsep}{0pt}
  \item Return only confident same-entity matches as id pairs: \texttt{[shot\_a\_id, shot\_b\_id]}.
  \item Do not include explanations.
  \item Use strict JSON only:
\end{itemize}
\begin{verbatim}
{
  "matches": [[shot_a_id, shot_b_id], ...]
}
\end{verbatim}
If no confident match exists, return \texttt{\{"matches": []\}}.
\end{promptbox}

\end{document}